\pdfoutput=1

\documentclass[11pt]{article}

\usepackage[]{acl}

\usepackage{multirow}
\usepackage{times}
\usepackage{latexsym}
\usepackage{soul}
\usepackage{url}
\usepackage{bm}
\usepackage{hyperref}
\usepackage[utf8]{inputenc}
\usepackage{graphicx}
\usepackage{amsmath}
\usepackage{mathrsfs}
\usepackage{amsthm}
\usepackage{booktabs}
\usepackage{algorithm}
\usepackage{algorithmic}
\usepackage{booktabs}
\usepackage{amssymb}
\usepackage{color}
\usepackage{multicol}
\usepackage{subfigure}
\graphicspath{{Figure/}}
\usepackage[T1]{fontenc}

\usepackage[utf8]{inputenc}

\usepackage{microtype}

\title{Derivative-Free Optimization for Low-Rank Adaptation in Large Language Models}

\author{Feihu Jin$^{1,2}$, Yifan Liu$^{1,2}$ \and Ying Tan$^{1,2,3,4}$ \thanks{~ Corresponding Author}\\
  $^1$ School of Intelligence Science and Technology, Peking University\\
  $^2$ Institute for Artificial Intelligence, Peking University \\
  $^3$ National Key Laboratory of General Artificial Intelligence\\ 
  $^4$ Key Laboratory of Machine Perceptron (MOE), Peking University\\
\text{fhjin@stu.pku.edu.cn}, \text{liuyifan731@163.com}, \text{ytan@pku.edu.cn}}

\begin{document}
\maketitle
\begin{abstract}
Parameter-efficient tuning methods such as LoRA could achieve comparable performance to model tuning by tuning a small portion of the parameters. However, substantial computational resources are still required, as this process involves calculating gradients and performing back-propagation throughout the model. Much effort has recently been devoted to utilizing the derivative-free optimization method to eschew the computation of gradients and showcase an augmented level of robustness in few-shot settings. In this paper, we prepend the low-rank modules into each self-attention layer of the model and employ two derivative-free optimization methods to optimize these low-rank modules at each layer alternately. Extensive results on various tasks and language models demonstrate that our proposed method achieves substantial improvement and exhibits clear advantages in memory usage and convergence speed compared to existing gradient-based parameter-efficient tuning and derivative-free optimization methods in few-shot settings\footnote{We have released our code in: \url{https://github.com/stan-anony/derivative_free_lora_rank}}. 
\end{abstract}

\section{Introduction}
In recent years, there has been a rapid development in large language models (LLMs) \cite{radford2019language, Brown2020LanguageMA, OpenAI2023GPT4TR, DBLP:journals/corr/abs-2302-13971},
\begin{figure}[ht!]
      \centering
    \includegraphics[scale=0.41]{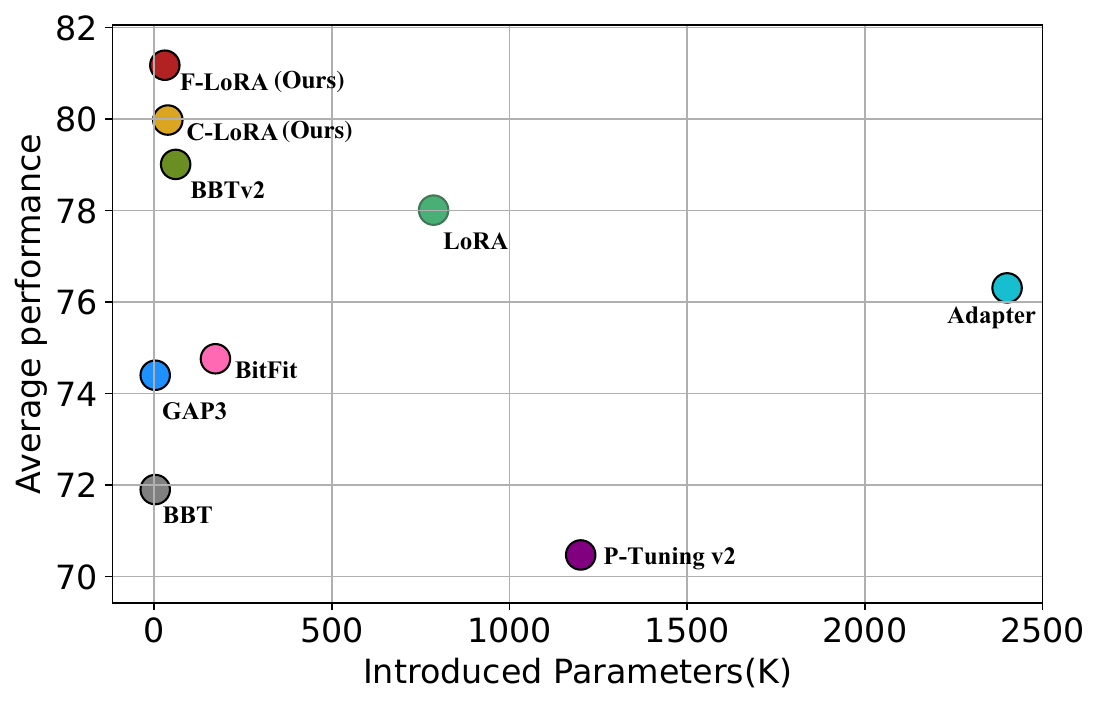}
    \caption{The results of our proposed methods F-LoRA and C-LoRA compared to gradient-based and gradient-free methods on average performance over seven language understanding tasks. We evaluate all the methods on RoBERTa-large.}
    \label{results}
\end{figure}
which have showcased impressive capabilities across various natural language processing tasks. However, the sheer number of parameters of large models leads to a linear increase in tuning cost and poses challenges for fine-tuning on common hardware. To this end, parameter-efficient tuning methods \cite{he2022towards,Houlsby2019ParameterEfficientTL,chen-etal-2022-inducer,chen-etal-2022-empowering} have emerged as a solution that could achieve comparable performance to full fine-tuning while only tuning a small portion of the parameters in large language models. Although these methods can reduce GPU memory requirements by approximately 30\% \cite{sung2022lst}, tuning a small subset of parameters still involves the computation of gradients and back-propagation, which presents challenges for utilizing and deploying large language models. Currently, the prevailing approach to harness the power of large language models is through in-context learning, treating the model as a service \cite{Brown2020LanguageMA}. The approach only involves forward computation and requires designing appropriate prompts or demonstrations without updating model parameters. However, in-context learning demands a meticulous selection of prompts and demonstrations, and the model's performance relies entirely on the chosen prompts and demonstrations \cite{gao2020making}.

Recently, much effort has been devoted to black-box tuning methods \cite{sun2022bbt,sun-etal-2022-bbtv2,DBLP:conf/ijcai/ZhaoWY23,Xu2023WizardLMEL,Oh2023BlackVIPBV}, which utilizes the derivative-free optimization method to optimize the introduced continuous prompts. Black-box tuning methods achieve comparable performance to parameter-efficient tuning methods and full fine-tuning in a few-shot setting without needing gradient computation and back-propagation through the entire LLM. However, it is acknowledged that training the prompt vectors in few-shot settings is prone to instability and exhibits slow convergence  \cite{lester2021power,li-liang-2021-prefix,Liu2021GPTUT}, making it challenging to generalize to large language models. 

In this paper, we propose a derivative-free optimization approach to address the challenges associated with the introduced low-rank modules in large language models that eschews the computation of gradients and showcases an augmented level of robustness in few-shot settings. Our method eliminates the need for gradient computation and back-propagation, resulting in improved stability and faster convergence compared to previous baselines. We prepend low-rank modules into each self-attention layer of the language model and initialize these modules by computing the mean and variance of hidden states for each layer. To optimize the parameters of the low-rank modules, we employ two derivative-free optimization methods. Recognizing that directly optimizing all low-rank modules of each layer using derivative-free methods in a high-dimensional space may slow down convergence, we adopt a divide-and-conquer strategy. The strategy entails optimizing the low-rank modules of each layer separately. To enable the optimization of the low-rank modules through derivative-free methods, we introduce a linear mapping matrix. The matrix maps the parameters obtained after derivative-free optimization to the desired low-rank modules at each layer. We initialize the linear mapping matrix based on normal distributions, with standard deviations related to the hidden states of each layer. 

We conduct comprehensive experiments on RoBERTa-large \cite{Delobelle2020RobBERTAD}, GPT2-large, and GPT2-XL \cite{radford2019language} to assess the effectiveness of our method. The results demonstrate that our proposed method has a significant improvement on average across seven natural language understanding tasks in a few-shot setting. As shown in the Figure \ref{results}, our proposed approach achieves substantial improvement compared to existing gradient-based parameter-efficient methods (e.g., Adapter tuning, LoRA, P-Tuning v2, and BitFit) and derivative-free optimization methods (e.g., BBT, GAP3, and BBTv2) on RoBERTa-large. Additionally, our proposed method demonstrates superior performance, clear advantages regarding GPU memory usage, and faster model convergence speed compared to existing derivative-free optimization methods in larger models.
\section{Preliminaries}
\subsection{Derivative-free Optimization} 
Derivative-free optimization (DFO) algorithms \cite{DBLP:journals/jmlr/WierstraSGSPS14,DBLP:journals/jgo/RiosS13,DBLP:conf/ijcai/QianHY16} are capable of tackling complex problems without relying on the back-propagation. Typically, these DFO algorithms employ a sampling-and-updating framework to enhance the solution iteratively. These algorithms have broad applications spanning various fields, from automatic machine learning \cite{NIPS2012_05311655} to reinforcement learning \cite{DBLP:journals/corr/SalimansHCS17,Bai_Zhang_Tao_Wu_Wang_Xu_2023} and objective detection \cite{DBLP:conf/cvpr/ZhangSVPL15}. Representative DFO algorithms include CMA-ES (Covariance Matrix Adaptation Evolution Strategy) \cite{6790628}, Fireworks algorithm \cite{8239679,9504974,9185563}, Genetic algorithms \cite{DBLP:books/daglib/0019083}, among others. CMA-ES is a widely adopted evolutionary algorithm for nonlinear and non-convex continuous optimization. It generates new potential solutions by sampling from a multivariate normal distribution model at each iteration. Besides, we have a Fireworks algorithm (FWA) based on simulating the explosion process of fireworks, introducing randomness and diversity to aid in escaping local minima and conducting a more comprehensive search of the problem space. FWA presents a new search manner that searches the potential space by a stochastic explosion process within a local space. Based on CMA-ES and FWA, we propose two derivative-free optimization methods for low-rank adaptation: C-LoRA and F-LoRA. We detail the optimization processes of the two gradient-free optimization methods in Appendix \ref{cmaes} and \ref{fwa}.
\subsection{Black-Box-Tuning} 
Common language understanding tasks can be formulated as classification tasks by incorporating task-specific prompts and a few labeled samples or by carefully engineering prompts and verbalizers \cite{Brown2020LanguageMA,schick-schutze-2021-exploiting}. For example, an input sentence combined with template $P$ that includes a <MASK> token can be represented as $X = \{x_1, x_2, \cdots, x_L, P,\text{<MASK>}.\}$, and $L$ corresponds to the length of the input sentence. When $X$ is fed into model $f$, the model can determine whether the corresponding label token of class $Y$ (e.g., "Yes" or "No") is more appropriate to replace the <MASK> token \cite{gao2020making}. Prompt tuning \cite{lester2021power} and P-tuning \cite{Liu2021GPTUT} insert continuous prompt vectors $\bm{p}\in \mathbb{R}^D$ into the input $X$ at the embedding layer, where the objective can be formulated as follows:
\begin{equation}
\begin{aligned}
\bm{p^{\star}} = \mathop{\arg\min}_{\bm{p} \in \Theta} \mathcal{L}(f(\bm{p};X),Y)
\end{aligned}
\end{equation}
where $\mathcal{L}$ is the loss function, $\Theta$ is the search space, and $\bm{p^{\star}}$ is the optimal prompt vector after a gradient-based optimization through the model of $f$. 

Recently, a gradient-free prompt tuning method, BBT \cite{sun2022bbt}, was proposed to learn the continuous prompts without back-propagation through the model. BBT utilizes derivative-free optimization algorithms to optimize the continuous prompt as follows:
\begin{equation}
\begin{aligned}
\bm{z^{\star}} = \mathop{\arg\min}_{\bm{z} \in \mathcal{Z}} \mathcal{L}(f(\bm{Az};X),Y)
\end{aligned}
\end{equation}
where $\bm{A}\in \mathbb{R}^{D\times d}$ is the random projection matrix, $\mathcal{Z}$ is the search space, and $\bm{z}\in \mathbb{R}^d$ is a low-dimensional subspace. BBT adopts the Covariance Matrix Adaptation Evolution Strategy (CMA-ES) \cite{6790628} to obtain the optimal prompt $\bm{z}^{\star}$. Inspired by the success of deep prompt tuning \cite{li-liang-2021-prefix,DBLP:journals/corr/abs-2110-07602}, BBTv2 \cite{sun-etal-2022-bbtv2} extends BBT by optimizing the deep prompt with derivative-free methods injected at every intermediate layer of the language model.
\section{Approach}
\begin{figure*}[t]
      \centering
    \includegraphics[scale=0.47]{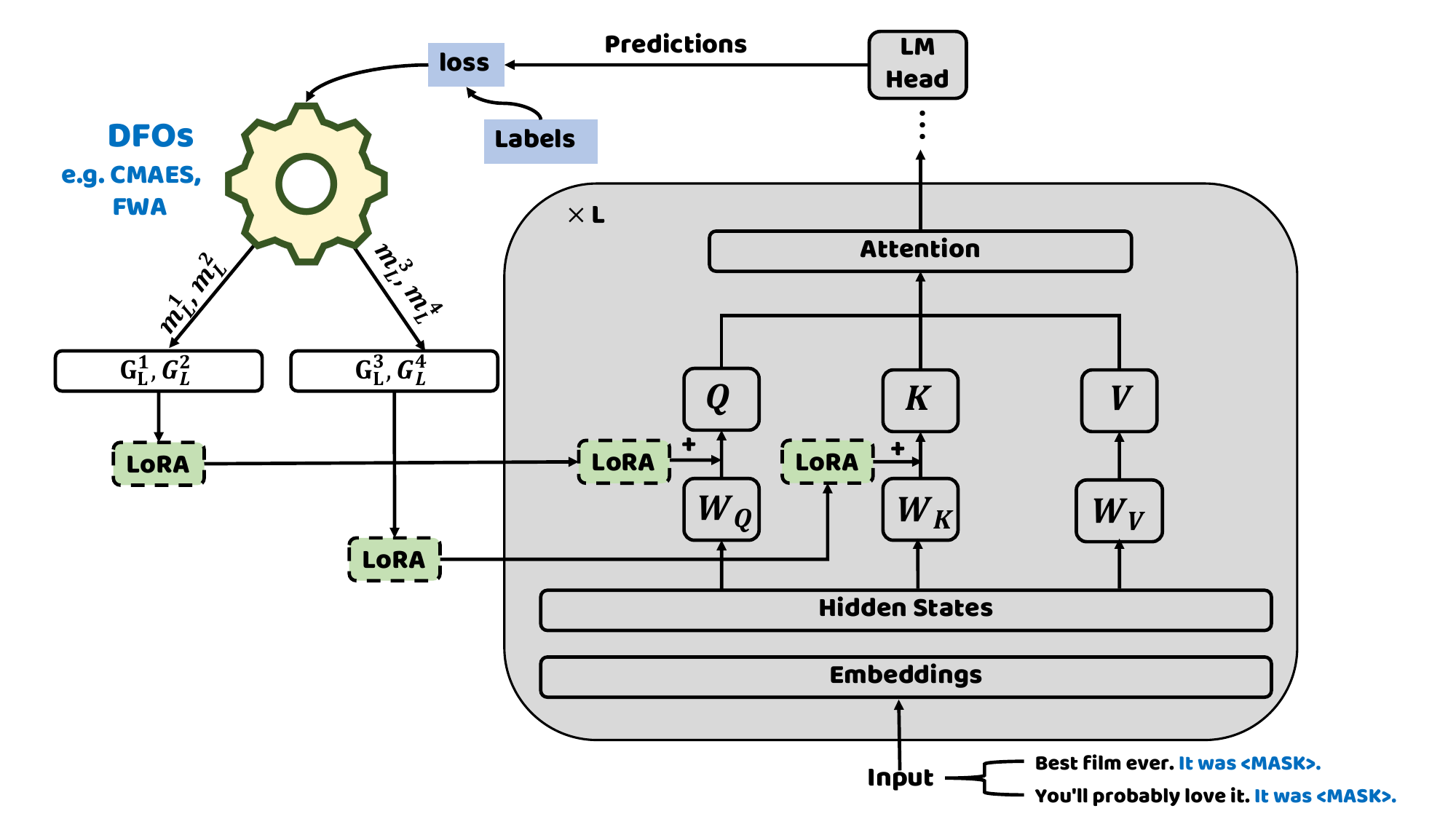}
    \caption{An illustration of derivative-free optimization for low-rank adaptation. We apply the low-rank matrices (green boxes) at the self-attention module of each layer and initialize them with model-specific normal distributions. We use two derivative-free methods (e.g., CMA-ES and Firework algorithm) to alternately optimize low-rank modules at the self-attention module of each layer.}
    \label{model}
\end{figure*}
The intrinsic dimensionality refers to the minimum dimension required to address high-dimensional optimization problems. It plays a crucial role in explaining the effectiveness of fine-tuning in language models \cite{li_id_2018_ICLR,aghajanyan-etal-2021-intrinsic}. Previous research on the intrinsic dimensionality of pre-trained language models has led to the development of approaches such as LoRA \cite{hu2022lora}. LoRA introduces down-sampling and up-sampling matrices at each layer of the Transformer and employs a low-rank decomposition optimization method to modify the original weight matrices within the self-attention modules. However, LoRA requires gradient back-propagation throughout the entire model, which may be computationally intensive when dealing with large language models. To further explore the potential benefits of combining low-rank adaptation optimization with derivative-free optimization, as depicted in Figure \ref{model}, we formally introduce our method, a derivative-free optimization for low-rank adaptation in large language models. 

Similar to the manual prompt learning family of models \cite{schick-schutze-2021-exploiting,schick-schutze-2021-just,gao2020making}, we first design our model to be close to the pre-training stage (e.g., keeping consistent objective function) by converting each input $X$ to masked language model (MLM) input, which contains a <MASK> token. Then, the model determines the corresponding verbalizers of class $Y$ to substitute for the <MASK> token. 

Building upon the success of Low-rank adaptation (LoRA) \cite{hu2022lora} and Black-Box prompt tuning (BBT) \cite{sun2022bbt,sun-etal-2022-bbtv2}, we incorporate low-rank matrices into the self-attention module of each layer in the pre-trained language model. We optimize the introduced low-rank matrix parameters using derivative-free optimization methods. As shown in Figure \ref{model}, the LoRA module consists of two low-rank matrices: $A \in \mathbb{R}^{r\times k}$ and $B\in \mathbb{R}^{D \times r}$. Here, $r$ represents the rank size, typically chosen as 2, 4, 8, 16, etc. For the weight matrices in the self-attention module $W\in \mathbb{R}^{D\times k}$ of the self-attention layer, the parameter updates are performed through matrix decomposition $W + \Delta W = W + BA$. During training, the parameters of $W$ are frozen and do not receive gradient updates, while the parameters of $A$ and $B$ are updated using gradient-free methods.

Considering that large language models have a low intrinsic dimensionality, we further optimize the parameters within a low-rank space using two gradient-free optimization methods. Figure \ref{model} illustrates this process. At the self-attention module of each layer in the language model, we optimize the vectors $\bm{m_L^1}\in \mathbb{R}^d$, $\bm{m_L^2}\in \mathbb{R}^d$, $\bm{m_L^3}\in \mathbb{R}^d$, and $\bm{m_L^4}\in \mathbb{R}^d$ using gradient-free optimizers such as FWA \cite{8239679,9504974,9185563} and CMA-ES \cite{6790628}, where $d$ is the intrinsic dimension. These optimized vectors are then projected into the low-rank space using specific random projection modules $\bm{G_L^1}\in \mathbb{R}^{r\times k\times d}$, $\bm{G_L^2}\in \mathbb{R}^{D\times r\times d}$, $\bm{G_L^3}\in \mathbb{R}^{r\times k\times d}$, and $\bm{G_L^4}\in \mathbb{R}^{D\times r\times d}$. The computation can be expressed as follows:
\begin{equation}
\begin{aligned}
A_Q &= \bm{G_L^1}\bm{m_L^1}\\
B_Q &= \bm{G_L^2}\bm{m_L^2}\\
A_K &= \bm{G_L^3}\bm{m_L^3}\\
B_K &= \bm{G_L^4}\bm{m_L^4}
\end{aligned}
\end{equation}
After obtaining $A_Q\in \mathbb{R}^{r\times k}$, $B_Q\in \mathbb{R}^{D\times r}$, $A_K\in \mathbb{R}^{r\times k}$, and $B_K\in \mathbb{R}^{D\times r}$ as low-rank matrices, we explore the weight matrices in self-attention modules with the application of CMA-ES for LoRA in Appendix \ref{sec_d} and update the weight matrices $W_Q$ and $W_K$ as follows:
\begin{equation}
\begin{aligned}
W_Q &= W_Q + B_Q A_Q \\
W_K &= W_K + B_K A_K
\end{aligned}
\end{equation}

This process is performed separately for each model layer, treating the optimization of the low-rank matrices concatenated to the self-attention layer as a subproblem optimization process. Inspired by the divide-and-conquer approach, we employ a gradient-free optimization strategy for the introduced parameters across the entire model.
\begin{equation}
\begin{aligned}
\bm{m_L^{i\star}} = \mathop{\arg\min}_{\bm{m_L^i} \in \mathcal{M}}\mathcal{L}(f(\bm{G_L^i m_L^i};X),Y)
\end{aligned}
\end{equation}
where $\mathcal{M}$ is the search space, $\mathcal{L}$ is the loss function, and $\bm{m_L^{i\star}}$ is the optimal vector after a derivative-free optimization through the model $f$. A detailed description is shown in Algorithm \ref{alg1}. 

The initialization of the modules $\bm{G_L^1}$, $\bm{G_L^2}$, $\bm{G_L^3}$, and $\bm{G_L^4}$ plays a crucial role in the performance of the model. We analyze two different initialization methods: random initialization with the normal distribution (e.g., initially set to $\mathcal{N} (0, 0.5)$) and initialization with the distribution of the hidden states at each layer of the language model similar to BBTv2 \cite{sun-etal-2022-bbtv2}. We show the details in Appendix \ref{initialize_g}. In section \ref{g_module}, our findings reveal that random initialization with the normal distribution leads to a slight decline in the performance of the language model and slows down the convergence. Therefore, we initialize the modules $\bm{G_L^1}$, $\bm{G_L^2}$, $\bm{G_L^3}$, and $\bm{G_L^4}$ using the distribution of the hidden states at each layer of the language model. This initialization strategy helps maintain the performance and convergence speed of the model, leading to better results.
\begin{algorithm}[ht]
	\caption{DFOs for Low-Rank Adaptation}
	\label{alg1}
	\begin{algorithmic}[1]
		\REQUIRE \ $L$-layer language model $f$, \\  \quad\quad\quad Budget of API calls: $\mathcal{B}$, \\ \quad\quad\quad Optimizers: $\{\mathcal{O}_i\}_{i=1}^L$,\\ \quad\quad\quad
        Loss function $\mathcal{L}$
		\STATE Hidden variable: $\bm{m_L^i}, i=1,2,3,4$
            \STATE Random projections: $\bm{G_L^i}, i=1,2,3,4$
            \STATE Low-rank modules of each layer: $A$ and $B$
        \REPEAT
        \FOR{each hidden layer}
            \STATE Evaluate: $loss = \mathcal{L}(f(\bm{G_L^i}\bm{m_L^i}))$ 
            \STATE Update $\bm{m_L^i}$ by DFOs: $\mathcal{O}_L^i(\bm{m_L^i}, loss)$
            \STATE Update the Low-rank modules $A$ and $B$ through $\bm{G_L^i}\bm{m_L^i}$ while keep $\bm{G_L^i}$ frozen
        \ENDFOR
        \UNTIL $\mathcal{B}/L$ times $f$ call
	\end{algorithmic}  
\end{algorithm}

\begin{table*}[ht]
\centering
\resizebox{\textwidth}{!}{
\renewcommand{\arraystretch}{1.4}
\begin{tabular}{@{}lrccccccc|c@{}}
\toprule 
\textbf{Method} & \textbf{\#Params} & 
  \textbf{SST-2}&
  \textbf{Yelpp} &
  \textbf{AG’s News}&
  \textbf{DBPedia}&
  \textbf{MRPC} &
  \textbf{SNLI} &
  \textbf{RTE} &
\textbf{Avg.}
  \\ \midrule
& & & &  \textbf{Gradient-Based Methods} \\ \midrule
Model tuning  & \textbf{355M}  & 85.49(2.84)  &91.82(0.79)  & 86.36(1.85)        & 97.98(0.14)         & 77.35(5.70) & 54.64(5.29)          & 58.60(6.21)  &78.88 \\
Adapter Tuning & \textbf{2.4M}  & 83.91(2.90) & 90.99(2.86) & 86.01(2.18)        & 97.99(0.07)          & 69.20(3.58) & 57.46(6.63)          & 48.62(4.74)& 76.31  \\
BitFit & \textbf{172K} & 81.19(6.08)          & 88.63(6.69)& 86.83(0.62)        & 94.42(0.94)          & 66.26(6.81)& 53.42(10.63)          & 52.59(5.31)  &74.76   \\ 
LoRA & \textbf{786K}  & 88.49(2.90)         & 90.21(4.00) & 87.09(0.85)        & 97.86(0.17)          & 72.14(2.23) & 61.03(8.55)          & 49.22(5.12)   &78.01\\
Prompt Tuning & \textbf{50K} & 68.23(3.78)         & 61.02(6.65) & 84.81(0.66)        & 87.75(1.48)          &51.61(8.67) & 36.13(1.51)          & 54.69(3.79) &63.46 \\
P-Tuning v2  &\textbf{1.2M} & 64.33(3.05)          & 92.63(1.39) & 83.46(1.01)        & 97.05(0.41)          &68.14(3.89) & 36.89(0.79)          & 50.78(2.28)  & 70.47\\ 
\midrule
& & & &  \textbf{Gradient-Free Methods} \\
\midrule
Manual Prompt & \textbf{0}  & 79.82          & 89.65 & 76.96        & 41.33         & 67.40 & 31.11          & 51.62   &62.56 \\
In-Context Learning  & \textbf{0}  & 79.79(3.06)          & 85.38(3.92) &62.21(13.46)       & 34.83(7.59)          & 45.81(6.67) & 47.11(0.63)          & 60.36(1.56)    &59.36\\
Feature-MLP & \textbf{1M}  & 64.80(1.78)          & 79.20(2.26) & 70.77(0.67)       &87.78(0.61)          & 68.40(0.86) & 42.01(0.33)         & 53.43(1.57) &66.63  \\
Feature-BiLSTM & \textbf{17M}  &65.95(0.99)          & 74.68(0.10) & 77.28(2.83)        & 90.37(3.10)          & 71.55(7.10) & 46.02(0.38)          & 52.17(0.25)   &68.29 \\
GAP3  & \textbf{2.5K}  & 89.70(2.80)          & 93.00(2.30) & 83.20(3.20)        & 83.70(2.90)          & 70.20(4.50)  & 51.10(4.60)          &49.70(1.50)  & 74.40\\
BBT  & \textbf{2.5K} &89.56(0.25)         & 91.50(0.16) & 81.51(0.79)      & 79.99(2.95)          & 61.56(4.34) & 46.58(1.33)         & 52.59(2.21) & 71.90\\
BBTv2  & \textbf{60K}  & 90.33(1.73)          & 92.86(0.62)& 85.28(0.49)     & 93.64(0.68)          & 77.01(4.73) & 57.27(2.27)          & 56.68(3.32) &79.01
\\ \midrule
\textbf{C-LoRA}  & \textbf{38K} &90.70(1.30)         & 93.37(0.44) & 85.55(0.57)      & 93.70(0.88)          & \textbf{80.12(1.88)} & 59.11(1.89)         & 57.34(2.34) & 79.98\\
\textbf{F-LoRA}  & \textbf{38K}  & \textbf{91.56(0.87)}          & \textbf{94.84(0.57)}& \textbf{86.64(0.55)}     & \textbf{94.95(0.54)}          & 79.99(1.67) & \textbf{61.42(1.47)}         & \textbf{60.93(1.42)} &\textbf{81.48}
\\ \bottomrule
\end{tabular}
}
\caption{Performance of gradient-based and gradient-free methods on RoBERTa-large. We report average and standard deviation performance over five different seeds. Bold fonts indicate the best results on derivative-free methods.}
\label{table_roberta}
\end{table*}

\begin{table*}[ht]
\centering
\resizebox{\textwidth}{!}{%
\renewcommand{\arraystretch}{1.4}
\begin{tabular}{@{}lrccccccc|c@{}}
\toprule
\textbf{Method} & \textbf{\#Params} & 
  \textbf{SST-2}&
  \textbf{Yelpp} &
  \textbf{AG’s News}&
  \textbf{DBPedia}&
  \textbf{MRPC} &
  \textbf{SNLI} &
  \textbf{RTE} &
\textbf{Avg.}
  \\ \midrule
& & & &  \textbf{GPT2-large} \\ \midrule
BBT  & \textbf{2.5K} &75.53(1.98)         & 80.75(0.53) & 77.63(1.89)      & 77.46(0.69)          & 65.56(2.34) & 32.28(2.43)         & 52.44(3.32) & 65.95\\
BBTv2  & \textbf{60K}  & 83.72(3.05)         & 85.46(2.33)& \textbf{79.96(0.75)}     & 91.36(0.73)          & 75.92(3.42) & 35.79(1.47)          & 55.78(1.42) &72.57
\\\midrule
\textbf{C-LoRA}  & \textbf{38K} &84.86(2.02)         & 87.75(0.91) & 79.46(0.87)    & 92.23(0.57)          & 76.20(1.22) & 37.08(0.99)         & 57.40(1.67) & 73.56\\
\textbf{F-LoRA}  & \textbf{38K}  & \textbf{85.88(1.51)}          & \textbf{88.52(0.55)}& 79.21(3.01)     & \textbf{92.44(0.76)}          & \textbf{76.44(0.67)} & \textbf{37.56(0.33)}         & \textbf{57.88(1.38)} &\textbf{74.00}
\\\midrule
& & & &  \textbf{GPT2-XL} \\
\midrule
BBT  & \textbf{5K} &78.56(1.46)         & 82.34(2.46) & 78.21(2.46)      & 79.76(3.65)          & 68.44(2.74) & 33.42(2.63)         & 54.08(2.49) & 67.83\\
BBTv2  & \textbf{120K}  & 85.86(2.45)          & 86.43(0.53)& 79.10(3.20)     & 92.14(1.35)          & 76.03(2.22) & 35.98(1.12)          & 55.23(2.47) &72.97
\\\midrule
\textbf{C-LoRA}  & \textbf{76K} &86.96(1.50)         & \textbf{88.70(0.72)} & 79.55(2.20)      & 92.83(0.55)          & 77.45(1.67) & 38.13(1.22)         & \textbf{58.65(1.44)} &74.61 \\
\textbf{F-LoRA}  & \textbf{76K}  & \textbf{87.33(1.67)}          & 88.47(1.24)& \textbf{79.84(1.88)}     & \textbf{93.87(0.94)}          & \textbf{78.09(1.26)} & \textbf{38.89(1.34)}         & 58.45(0.97)
&\textbf{74.99}
\\ \bottomrule
\end{tabular}%
}

\caption{Performance of gradient-free methods on GPT2-large. We report average  and standard deviation performance over five different seeds. Bold fonts indicate the best results.}
\label{table_gpt}
\end{table*}
\section{Experiments}
This section details the experimental results of several natural language understanding (NLU) tasks. The results demonstrate that our proposed method outperforms the current gradient-based and gradient-free methods on several tasks in a few-shot setting.
\subsection{Dataset Statistics}
We conduct extensive experiments on seven standard NLU datasets, which cover a range of tasks, including natural language inference, paraphrase identification, sentiment analysis, and topic classification. The detailed statistics of these datasets are shown in Appendix \ref{dataset}.
\subsection{Experimental Settings}
\noindent \textbf{Datasets} \ The implementation of our method is based on HuggingFace \cite{Wolf2020TransformersSN} and Pytorch \cite{Paszke2019PyTorchAI}. Considering the incredible power of large language models in a few-shot setting \cite{Brown2020LanguageMA}, we conduct our experiments with the same procedure as \citet{zhang2021revisiting}, \citet{gu-etal-2022-ppt}, and \citet{sun-etal-2022-bbtv2} to construct the true few-shot learning settings \cite{perez2021true}. We sample
$n$ instances for each class $\mathcal{Y}$ from the original training set to form the true few-shot training set $\mathcal{D}_{\text{train}}$ and validation sets $\mathcal{D}_{\text{dev}}$, and ensure that $|\mathcal{D}_{\text{train}}| = |\mathcal{D}_{\text{dev}}|$. In particular, in our experiments, the size of the test sets is significantly larger than that of the training and validation sets.

\noindent \textbf{Hyperparameters} \ 
We use a default setting training with a population size of 20 and a budget of 6,000 API calls to all the tasks for CMA-ES and a default setting training with a population size of 5 and a budget of 6,000 API calls to all the tasks for FWA. We set the rank $r$ of the low-rank matrix to be 2 or 4. We train our proposed method on one NVIDIA 3090 with 24G of memory and report the accuracy for all datasets except the F1 score for MRPC dataset. For generating random projections, we use normal distributions with standard deviations initialized with the distribution of the hidden states at each layer of the language model.
\subsection{Baselines}
To ensure a fair comparison, we use RoBERTa-large \cite{Delobelle2020RobBERTAD} as the pre-trained model for both gradient-based and gradient-free methods in our experiments. Specifically, to verify the effectiveness of our method on large language models, we chose GPT2-XL \cite{radford2019language} as a large language model for gradient-free methods. We compare our proposed method (\textbf{C-LoRA}) and FW-LoRA (\textbf{F-LoRA}) with several baselines as follows:

\noindent\textbf{Gradient-based methods} For Gradient-based methods, we compare with (1) \textbf{Model tuning:} The vanilla transformer fine-tuning \cite{Delobelle2020RobBERTAD}. (2) \textbf{Adapter tuning:} Inserting a small task-specific module between the self-attention module (and the MLP module) and the subsequent residual connection at each Transformer layer \cite{Houlsby2019ParameterEfficientTL}. (3) \textbf{BitFit:} Tuning the biases of the pre-trained language model in our few-shot settings \cite{ben-zaken-etal-2022-bitfit}. (4) \textbf{LoRA:} Merging the low-rank and trainable matrices with the frozen weights at each layer of the Transformer \cite{hu2022lora}. (5) \textbf{P-Tuning v2:} Appending trainable continuous prompt vectors at each layer of the Transformer \cite{DBLP:journals/corr/abs-2110-07602}. (6) \textbf{Prompt tuning:} Appending trainable continuous prompt vectors at embedding layer of the Transformer \cite{lester2021power}.

\noindent\textbf{Gradient-free methods} For Gradient-free methods, we compare with (1) \textbf{Manual Prompt:} Using the templates and label words to conduct zero-shot evaluation \cite{gao2020making}. (2) \textbf{In-context learning:} Selecting up several training samples and concatenating them with the input texts \cite{Brown2020LanguageMA}. (3) \textbf{Feature-MLP} and (4) \textbf{Feature-BiLSTM:} Training a MLP/BiLSTM classifier on the features extracted by the language model  \cite{peters-etal-2019-tune}. (5) \textbf{GAP3:} Black-box prompt tuning with genetic algorithm  \cite{DBLP:conf/ijcai/ZhaoWY23}. (6) \textbf{BBT:} Black-box prompt tuning with CMA-ES algorithm. (7) \textbf{BBTv2:} Black-box deep prompt tuning with CMA-ES algorithm.

\subsection{Main Results}
\noindent\textbf{Overall Comparison on RoBERTa Model.} As illustrated in Table \ref{table_roberta}, we demonstrate the experimental results of our proposed methods and the baselines across seven datasets. We observe that the proposed methods outperform gradient-based and gradient-free methods, exhibiting varying levels of improvement across different NLP tasks in a few-shot settings, on average. Specifically, for gradient-based methods, the proposed method improves the performance compared to Model tuning, Adapter tuning, BitFit, LoRA, and P-Tuning v2 by an average of 2.60, 5.17, 6.72, 3.47, and 11.01 points, respectively, across the seven NLU datasets. Furthermore, compared to these parameter-efficient fine-tuning methods, we introduce fewer parameters, yielding superior model performance. For gradient-free methods, the proposed method improves the performance compared to Manual Prompt, In-Context-Learning, GAP3, BBT, and BBTv2 by an average of 18.92, 22.12, 7.05, 9.58, and 2.47 points, respectively, across the seven NLU datasets. Our proposed methods perform better than gradient-free methods across all datasets (e.g., especially on RTE, MRPC, and SST-2), demonstrating the effectiveness of our methods. It is important to emphasize that the primary distinction between LoRA and our proposed method lies in the optimization algorithm employed. LoRA utilizes gradient descent (Adam optimizer), whereas our method employs the DFOs algorithm (C-LoRA and F-LoRA). Our experimental results indicate that gradient-based optimization may lead to overfitting on limited training data, while DFOs, with their exploration mechanism, tend to discover more effective solutions.

\noindent\textbf{Overall Comparison on GPTs Models.}
To evaluate the efficacy of our proposed approach on larger models, we conduct experiments on the GPT2-large and GPT2-XL models, as illustrated in Table \ref{table_gpt}. We show the performance of the two gradient-free optimization methods across models of varying sizes compared to other baselines. Specifically, for GPT2-large, the proposed method improves the performance compared to BBT and BBTv2 by 8.05 and 1.43 points, respectively, on average. It even outperforms BBTv2 across 6/7 datasets (e.g., SST2, Yelpp, DBPedia, MRPC, SNLI, and RTE). For the larger model GPT2-XL, the proposed method improves the performance by 7.16 and 2.02 points on average, respectively, and even outperforms BBTv2 across all datasets. Moreover, as the parameters of the model continue to increase, it can be found that the proposed method is still effective and performs better than other gradient-free optimization methods when fewer parameters are introduced, demonstrating that the proposed method can generalize to more large language models.

\section{Analysis}
\subsection{Memory Usage and Training Time}
Table \ref{table_time} compares our proposed methods (C-LoRA and F-LoRA) with BBTv2 regarding memory usage and training time on SST2, AG's News, and MRPC datasets. The experiments are conducted with a batch size of 16 and a sequence length 512 in GPT2-XL. To monitor GPU memory usage during training, we use Nvidia-smi. Our proposed methods demonstrate superior performance with reduced GPU memory consumption compared to BBTv2. To ensure a fair comparison of training time, we utilize a single NVIDIA 3090 GPU with 24GB of memory, implementing early stopping if the development accuracy does not improve after 1,500 steps. Our methods exhibit faster convergence than BBTv2, achieving improvements of 8.7 minutes on SST2, 15.4 minutes on AG's News, and 7.7 minutes on MRPC. This indicates that our approach has the potential to be applied to large language models, offering parameter-efficient, memory-efficient, and faster convergence.
\begin{table*}[]
\centering
\resizebox{\textwidth}{!}{
\renewcommand{\arraystretch}{1.3}
\begin{tabular}{@{}l|ccc|ccc|ccc@{}}
\toprule
\multirow{2}{*}{\textbf{Datasets/Methods}}
& \multicolumn{3}{c| }{SST2}                    & \multicolumn{3}{c|}{AG's News}                           
& \multicolumn{3}{c}{MRPC}
  \\                    & BBTv2          & C-LoRA          & F-LoRA          & BBTv2          & C-LoRA          & F-LoRA    & BBTv2          & C-LoRA          & F-LoRA          \\ \midrule
Accuracy (\%)                                           & 85.86          & 86.96            & 87.33          & 79.10          & 79.55          & 79.84  & 76.03          & 77.45          & 78.09         \\ \midrule
Memory Usage (MB)                                      & 12698          & 12044           & 12044           & 17838          & 17037           & 17037   & 13388   & 12780          & 12780       \\ \midrule
Training Time (mins)                                                         & 20.8          & 16.4           & 12.1          & 35.8          & 25.7           & 20.4         & 22.6         & 18.5          & 14.9  \\ \bottomrule
\end{tabular}
}
\caption{Comparison of BBTv2, C-LoRA, and F-LoRA on accuracy, memory usage and training time on a single NVIDIA 3090 GPU with 24GB of memory. Batch sizes are 16 and sequence lengths are 512.}
\label{table_time}
\end{table*}

\subsection{Effect of Subspace Dimensionality}
\begin{figure}[t]
\centering
\subfigure{
\begin{minipage}[t]{0.45\linewidth}
\centering
\includegraphics[scale=0.3]{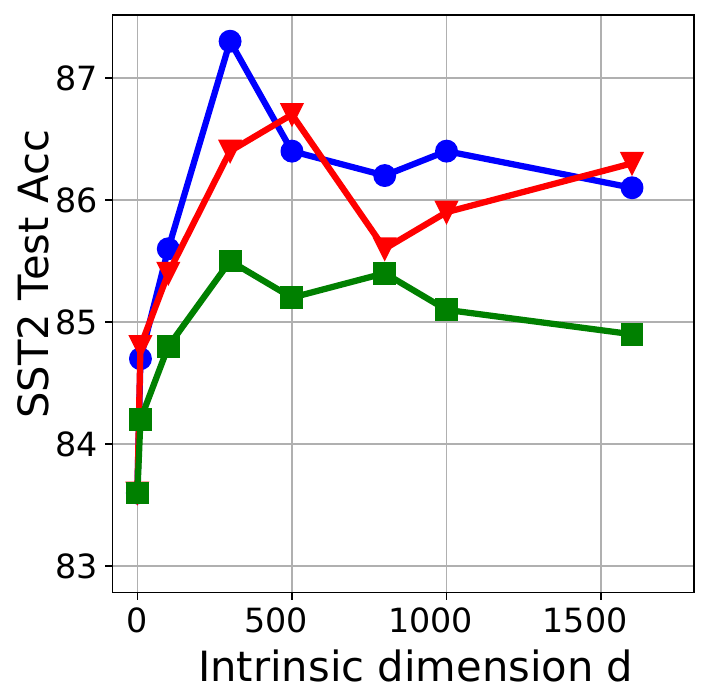}
\label{sst2_dim}
\end{minipage}%
}%
\subfigure{
\begin{minipage}[t]{0.45\linewidth}
\centering
\includegraphics[scale=0.3]{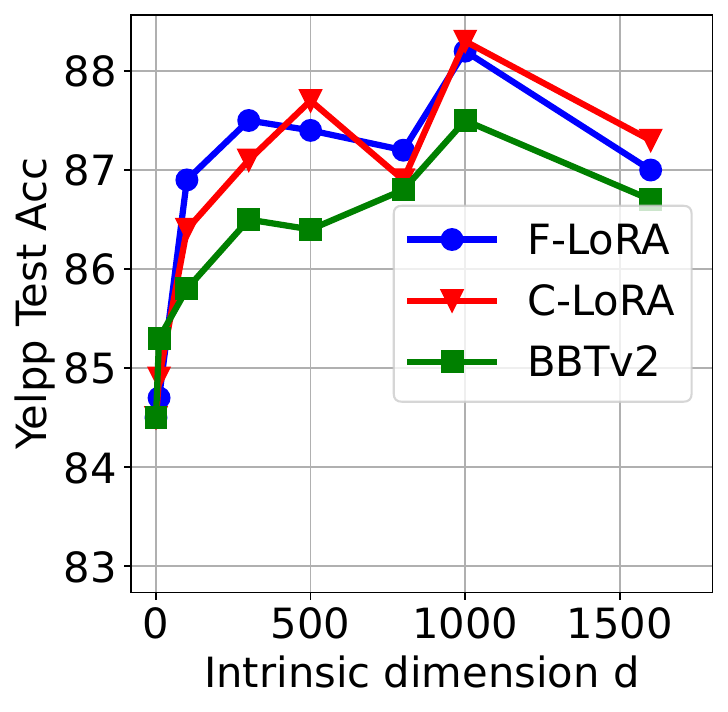}
\label{yelpp_dim}
\end{minipage}%
}
\caption{The results of different dimensions on the SST2 and SNLI datasets with GPT2-XL model.}
\label{figure_dim}
\end{figure}

In exploring the impact of subspace dimensionality on our proposed method, we employ the GPT2-XL model and conduct experiments on the SST2 and SNLI datasets, as illustrated in Figure \ref{figure_dim}. We explore the dimensionality ranges from 10 to 1600 using BBTv2, C-LoRA, and F-LoRA while maintaining a consistent batch size and population size. As observed in Figure \ref{figure_dim}, our proposed method consistently outperforms BBTv2 with the increasing subspace dimensionality. Simultaneously, we note that the performance improvement of the model gradually stabilizes when the subspace dimension $d>500$. Considering that increasing the dimension $d$ may cost much training time for gradient-free algorithms, we keep the range of $d$ between 500 and 1600. 
\subsection{Effect of Low-Rank $r$}
\begin{figure}[t]
\centering
\subfigure{
\begin{minipage}[t]{0.45\linewidth}
\centering
\includegraphics[scale=0.3]{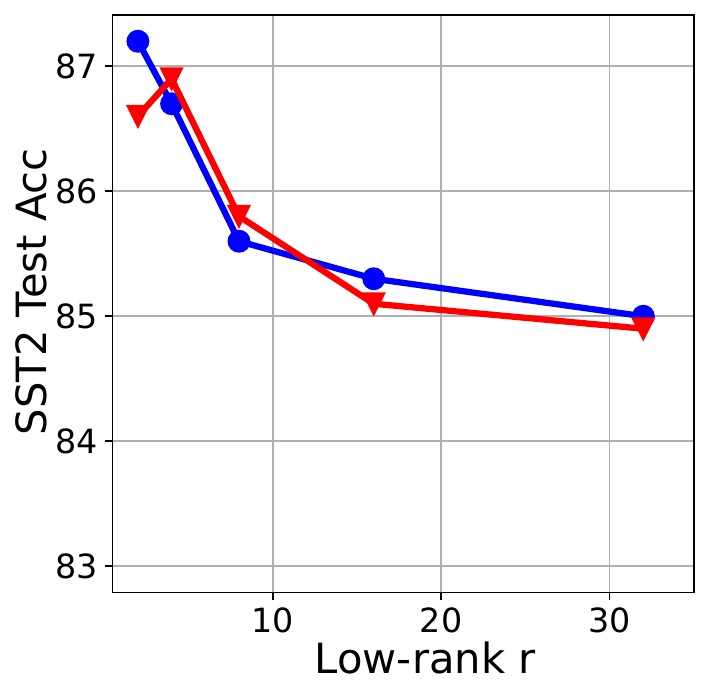}
\label{sst2_lora}
\end{minipage}%
}%
\subfigure{
\begin{minipage}[t]{0.45\linewidth}
\centering
\includegraphics[scale=0.3]{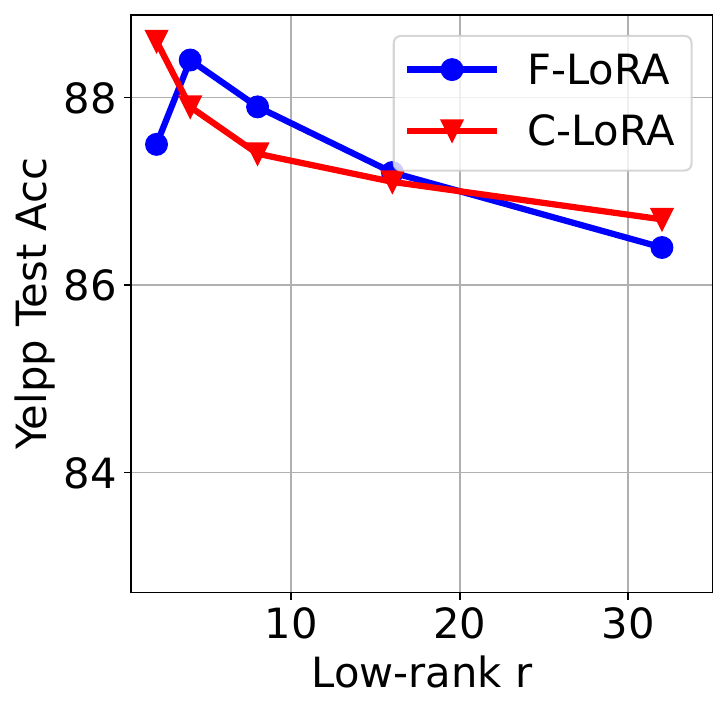}
\label{yelpp_lora}
\end{minipage}%
}
\caption{The results of different low-rank $r$ on the SST2 and Yelpp datasets with GPT2-XL model.}
\label{figure_lora}
\end{figure}

Considering the impact of the low-rank $r$ on the performance of our proposed method, we conduct experiments on SST2 and Yelpp datasets with the GPT2-XL model to analyze the importance of $r$. As indicated in Figure \ref{figure_lora}, we observe that as the low rank $r$ increases, the performance of our two proposed methods gradually decreases on the model. This suggests that when optimizing with gradient-free methods, the model does not require optimization in high dimensions, and achieving good results only requires optimization in a low rank $r$. In our experiments, we choose $r=2$ or 
$r=4$, allowing the model to achieve good results by introducing very few parameters. 

\begin{table}[!htbp]
\centering
\begin{tabular}{@{}lccccccc@{}}
\toprule
Methods & Yelpp (Acc.) & RTE (Acc.)\\ \midrule
RI  &  82.22(2.56)  & 52.71(3.11) \\
RI+DFO & 87.12(2.41)  & 57.10(2.69)\\
RIL  &   83.42(1.34) & 53.01(2.11)  \\
RIL+DFO & 88.70(0.72) & 58.65(1.44)
\\ \bottomrule
\end{tabular}
\caption{Test accuracy on Yelpp and RTE with GPT2-XL on different types of initialization of module $G$. 'RI' denotes random initialization with the normal distribution. 'RIL' denotes random initialization with the distribution of the hidden states at each layer of language model.}
\label{module_g}
\end{table}
\subsection{Effect of Initialization of Module $G$}\label{g_module}
In Table \ref{module_g}, we analyze two different initialization methods for module $G$ using DFO (C-LoRA) on Yelpp and RTE datasets: one involves random initialization with a normal distribution, and the other utilizes the distribution of hidden states at each layer of the language model for initialization. Experimental analysis reveals a significant performance degradation with the random initialization while initializing based on the distribution of the language model's hidden states further mitigates this phenomenon. Additionally, we observe that both the application of DFO (C-LoRA) and these projection matrices yield gains, with DFO making a more pronounced contribution. This suggests the effectiveness of gradient-free optimization methods in optimizing low-rank matrices.

\section{Related Work} 
\noindent\textbf{Efficient Few-shot Learners} 
\citet{peters-etal-2019-tune} and \citet{Dodge2020FineTuningPL} show that fine-tuning all parameters of language models in few-shot settings can be sub-optimal. As an efficient alternative, parameter-efficient tuning is a promising way of stimulating LLMs. We list mainstream parameter-efficient tuning models as follows: a) Adapter tuning methods learn the task-specific information \cite{Houlsby2019ParameterEfficientTL} by inserting small-scale task-specific modules within layers of Transformer and only tune the added adapters and layer normalization for model adaptation. b) Prefix tuning methods \cite{li-liang-2021-prefix,DBLP:journals/corr/abs-2110-07602} also introduce additional prompt vectors within layers of Transformer to learn task-specific information. Only the parameters of trainable prompt vectors are updated during training while keeping the model parameters frozen. Recently, several prompt tuning methods learn the instance-dependent information \cite{DBLP:journals/talip/JinL0Z23, DBLP:journals/corr/abs-2111-02643,DBLP:journals/corr/abs-2204-04497} by inserting the instance-dependent trainable continuous tokens to the input or hidden states of each Transformer layer. c) BitFit \cite{ben-zaken-etal-2022-bitfit}, a simple but effective method, only optimizes the bias terms inside the model while keeping other parameters frozen. d) LoRA \cite{hu2022lora} merges the low-rank and trainable matrices with the frozen weights at each layer of the Transformer. These parameter-efficient tuning methods still need gradient computation and backpropagation. However, our proposed method utilizes the gradient-free methods and optimizes the model in a memory-efficient and parameter-efficient way. 

\noindent\textbf{Gradient-free Optimization of LLMs} Gradient-free optimization methods have always been widely used in practice. It shows powerful potential in large language models. \citet{sung2022lst} introduced a method that eliminates the need for gradient updates by directly applying a pruned model to downstream tasks. \citet{DBLP:journals/corr/abs-2302-04870} and \citet{jin-etal-2023-parameter} propose an efficient transfer learning framework that can adapt large language models to downstream tasks without access to full model parameters. Recently, black-box tuning methods \cite{sun2022bbt,sun-etal-2022-bbtv2,Xu2023WizardLMEL,Oh2023BlackVIPBV} have employed Covariance Matrix Adaptation Evolution Strategy (CMA-ES) \cite{6790628,Hansen2003ReducingTT} to optimize continuous prompt vectors, bringing substantial benefits to the application of large models with low complexity. However, it is acknowledged that training the introduced prompt vectors is unstable and exhibits slower convergence \cite{lester2021power,li-liang-2021-prefix,Liu2021GPTUT}. Therefore, we propose gradient-free optimization for low-rank adaptation to overcome training instability and improve the speed of convergence.
\section{Conclusion}
In this work, we introduce a novel method for optimizing low-rank modules in large language models in a derivative-free way. The method involves integrating low-rank modules into each self-attention layer of the model and employing two derivative-free optimization methods to optimize these modules at each layer iteratively. Extensive experiments on different tasks and language models show that our proposed method demonstrates superior performance, lower GPU memory usage, and faster model convergence speed compared to existing derivative-free optimization methods in few-shot settings, suggesting that our method presents a promising direction for effectively and economically utilizing LLMs.
\section*{Limitations}
The proposed method is limited in its applicability to large models where obtaining weights is not feasible, as it requires modifying the specific model structure. Additionally, it is crucial to highlight that our method has only been validated in the context of language understanding tasks. Further exploration and investigation are necessary to assess its effectiveness in generation tasks.

\bibliography{anthology,custom}
\clearpage
\appendix
\section{Evolution Strategy}
\subsection{The CMA Evolution Strategy}\label{cmaes}
CMA-ES, short for Covariance Matrix Adaptation Evolution Strategy, is an evolutionary optimization algorithm designed explicitly for tackling optimization problems that are continuous and non-convex. One of the key distinguishing features of CMA-ES is its ability to dynamically adapt the population's covariance matrix, which facilitates efficient exploration and exploitation of the search space.
The algorithm maintains a distribution of candidate solutions generated based on a multivariate normal distribution. Through a series of iterations, CMA-ES continuously adjusts this distribution's mean and covariance matrix. This adaptive process enables the algorithm to explore and exploit promising regions within the search space.
During each iteration, CMA-ES refines the distribution by replacing less promising solutions with new candidate solutions generated from the current distribution. By dynamically adjusting the covariance matrix, the algorithm can focus on regions that show potential for improved solutions. This iterative process of adaptation and refinement allows CMA-ES to converge towards optimal or near-optimal solutions for the given optimization problem.
\subsection{The Fireworks Algorithm}\label{fwa}
The Fireworks Algorithm (FWA) utilizes a heuristic search approach based on two critical operations: the explosion and selection operations. During the explosion operation, multiple sparks are generated around existing fireworks within specified explosion amplitudes. These sparks serve as potential solutions for the next generation. Subsequently, the fireworks for the new generation are selected from these sparks.
We employ the loser-out tournament-based FWA (LoTFWA) to optimize the low-rank modules. In LoTFWA, fireworks compete, and the losers are forced to restart their search from a new location. This competitive mechanism relies on evaluating the fitness of each firework. If a firework's fitness fails to match the best fitness achieved so far, considering its current progress rate, it is considered a loser. The loser is then eliminated and reinitialized, as continuing its search process would be ineffective. This reinitialization step significantly reduces the likelihood of the algorithm becoming trapped in local minima.
\section{Patterns and Verbalizers}
\textbf{SST-2}
\begin{itemize}
\item{Pattern ["text", "It", "was", "<mask>", "."]}
\item{Verbalizers \{"0": "bad", "1": "great"\}}
\end{itemize}
\textbf{Yelpp}
\begin{itemize}
\item{Pattern ["text", "It", "was", "<mask>", "."]}
\item{Verbalizers \{"0": "bad", "1": "great"\}}
\end{itemize}
\textbf{AG's News}
\begin{itemize}
\item{Pattern ["<mask>", "News", "text", "."]}
\item{Verbalizers \{  "0": "World",
        "1": "Sports", "2": "Business", "3": "Tech"\}}
\end{itemize}
\textbf{DBPedia}
\begin{itemize}
\item{Pattern ["Category: <mask>", "text"]}
\item{Verbalizers \{ "0": "Company",
            "1": "Education",
            "2": "Artist",
            "3": "Athlete",
            "4": "Office",
            "5": "Transportation",
            "6": "Building",
            "7": "Natural",
            "8": "Village",
            "9": "Animal",
            "10": "Plant",
            "11": "Album",
            "12": "Film",
            "13": "Written"\}}
\end{itemize}
\textbf{SNLI}
\begin{itemize}
\item{Pattern ["text1", "?", "<mask>", ",", "text2"]}
\item{Verbalizers \{ "0": "Yes",
        "1": "Maybe",
        "2": "No"\}}
\end{itemize}
\textbf{RTE}
\begin{itemize}
\item{Pattern ["text1", "?", "<mask>", ",", "text2"]}
\item{Verbalizers \{"0": "Yes", "1": "No"\}}
\end{itemize}
\textbf{MRPC}
\begin{itemize}
\item{Pattern ["text1", "?", "<mask>", ",", "text2"]}
\item{Verbalizers \{  "0": "No",
        "1": "Yes"\}}
\end{itemize}

\section{Datasets}\label{dataset}
Table \ref{table-dataset} shows the statistics of datasets used in this work. Specifically, We evaluate our method on AG’s News \cite{NIPS2015_250cf8b5} and DBPedia \cite{NIPS2015_250cf8b5} for topic classification, RTE \cite{wang-etal-2018-glue} and SNLI \cite{bowman-etal-2015-large} for natural language inference, SST-2 \cite{socher-etal-2013-recursive} and Yelp \cite{NIPS2015_250cf8b5} for sentiment analysis, and MRPC \cite{dolan-brockett-2005-automatically} for semantic paraphrasing.
\begin{table*}[!htbp]
\centering
\begin{tabular}{@{}lcccc@{}}
\toprule
\textbf{Datasets} & |$\mathcal{Y}$| & |Train| & |Test|  & Task \\ \midrule
& & &  Single-sentence \\ \midrule 
SST-2   & 2 & 67k & 0.9k & Sentiment analysis  \\
Yelpp   & 2 & 560k & 38k &  Sentiment analysis \\
AG's News  & 4 & 120k & 7.6k &  Topic classification \\
DBPedia  & 14 & 560k & 70k & Topic classification \\
\midrule
& & & Sentence-pair \\ \midrule
SNLI    & 3 & 549k & 9.8k & Natural language inference \\
RTE   & 2 & 2.5k & 0.3k & Natural language inference \\
MRPC     & 2 & 3.7k & 0.4k & Semantic paraphrasing \\
\midrule
\end{tabular}
\caption{The datasets evaluated in this work. We sample
N ×|$\mathcal{Y}$| instances from the original training set to form the few-shot training and validation sets.  }
\label{table-dataset}
\end{table*}

\begin{table*}[!htbp]
\centering
\begin{tabular}{@{}lccccccc@{}}
\hline
Weight Type & $W_Q$ & $W_K$ & $W_V$  & $W_Q$, $W_K$ & $W_Q$, $W_V$ & $W_K$, $W_V$ & $W_Q$, $W_K$, $W_V$\\ 
Rank $r$  &  2 & 2 & 2 & 2 & 2 & 2 & 2 \\ \hline
Yelpp (Acc.)  &  87.64 & 86.45 & 85.66 & \textbf{88.70} & 87.34 & 86.48 & 86.95 \\
SNLI (Acc.) &  37.23 & 36.67 & 35.44 & \textbf{38.13} & 37.25 & 37.22 & 37.68
\\ \hline
\end{tabular}
\caption{Test accuracy on Yelpp and SNLI after applying the derivative-free optimized method on different types of weight matrices.}
\label{lora-weight}
\end{table*}
\section{Applying Gradient-free LoRA to Which Weight Matrices ?}\label{sec_d}
In our investigation of weight matrices in self-attention modules with the application of CMA-ES for LoRA, we have conducted experiments on the GPT2-XL model. Table \ref{lora-weight} shows that utilizing the derivative-free optimized LoRA only on the $W_Q$, $W_K$, and $W_V$ matrices leads to decreased performance. However, we have observed that simultaneous application of the derivative-free optimized LoRA to the $W_Q$ and $W_K$ matrices yields the best performance.
These findings underscore the significance of selecting the weight matrices for derivative-free optimization using the LoRA method. Through exploring various combinations, we can identify the most effective configuration for maximizing performance on the GPT2-XL model.

It is imperative to highlight a noteworthy distinction between our findings and the conventional LoRA approach. Traditionally, LoRA has demonstrated optimal outcomes by concatenating low-rank matrices on the $K$ and $V$ dimensions. However, our novel method, employing a gradient-free optimization approach, exhibits a predilection for achieving superior results by concatenating low-rank matrices on the $Q$ and $K$ dimensions. It is essential to underscore that this empirical observation is derived solely from experimental results and currently lacks a comprehensive theoretical analysis. We defer the in-depth examination of this intriguing phenomenon to future research endeavors.
\section{How to Initialize the Module $G$}\label{initialize_g}
Inspired by BBTv2, we first set the $\mu = 0$. Then we initialize module $G$ with a normal distribution using the standard deviation as follows:
\begin{equation}
\begin{aligned}
\sigma_m = \frac{\alpha\hat{\sigma}}{\sqrt{d}\sigma_z}
\end{aligned}
\end{equation}
where $\hat{\sigma}$ represents the observed standard deviation of hidden states, $\sigma_z$ denotes the standard deviation of the normal distribution maintained by DFOs, and $\alpha$ is a constant scalar used to stretch the distribution.
\end{document}